\documentclass[11pt]{article}

\usepackage{amsmath,amssymb,amsthm}
\usepackage{bm}
\usepackage{geometry}
\usepackage{booktabs}
\usepackage{enumitem}
\usepackage{hyperref}
\usepackage{natbib}
\usepackage{tikz}
\usetikzlibrary{arrows.meta}

\theoremstyle{plain}

\tikzset{
  varnode/.style={
    draw, rounded corners=2pt,
    minimum width=1.7cm, minimum height=0.5cm,
    font=\scriptsize\sffamily,
    fill=white, inner sep=2pt
  },
  directed/.style={-{Stealth[length=5pt]}, thick},
  undirected/.style={thick}
}
\newcommand{\radius}{3.0cm}
\newcommand{\circnode}[3]{%
  \node[varnode] (#1) at ({#2}:\radius) {#3};
}

\title{Fourier Feature Methods for Nonlinear Causal Discovery:\\
FFML Scoring, TRFF Scoring, and FFCI Testing in Mixed Data}

\author{%
  Joseph Ramsey \\
  Department of Philosophy \\
  Carnegie Mellon University \\
  Pittsburgh, PA 15213 \\
  \texttt{jdramsey@andrew.cmu.edu}
}

\date{}

\begin{document}
\maketitle

\begin{abstract}
Gaussian process (GP) marginal likelihood scores and kernel conditional independence tests are theoretically appealing for nonlinear causal discovery but computationally prohibitive at scale. We present three complementary RFF-based methods forming a practical toolkit for score-based, constraint-based, and hybrid causal discovery.

The Fourier Feature Marginal Likelihood (FFML) score approximates the exact GP marginal likelihood by replacing the $n x n$ kernel Gram matrix with a finite-dimensional feature representation, reducing cost to $O(nm^2 + m^3)$ while retaining the probabilistic interpretation and automatic complexity penalty of the exact score. FFML extends to mixed (continuous and discrete) parent sets via a product-kernel construction, with a Kronecker path for small discrete parent sets and a Hadamard-product path otherwise.

The Tetrad Random Fourier Feature (TRFF) score is a complementary BIC-style alternative using penalized Student-t regression with random Fourier features. TRFF offers robustness to heavy-tailed noise and faster runtime than FFML. Empirically, TRFF and FFML exhibit a complementary precision-recall profile: TRFF achieves higher precision while FFML achieves better recall and lower SHD overall.

The Fourier Feature Conditional Independence (FFCI) test is a fast nonparametric CI test for mixed data, using ridge residualization in feature space and a Frobenius-norm cross-covariance statistic approximated as a weighted sum of chi-squared variables.
Empirically, BOSS+FFML achieves the lowest SHD on nonlinear data, while BOSS+TRFF offers the highest precision. When run through PC-Max, FFCI and RCIT exhibit complementary precision-recall profiles: RCIT is more precise while FFCI achieves better recall and substantially lower SHD, at approximately twice the runtime.
\end{abstract}

% ==============================================================
\section{Background and Motivation}
% ==============================================================

Kernel-based methods provide a principled and flexible approach to
modeling nonlinear relationships in causal discovery.
They are attractive both for score-based algorithms, which evaluate
candidate parent sets via regression-style objectives, and for
constraint-based algorithms, which rely on conditional independence
(CI) testing.
In greedy score-based search procedures such as FGES \citep{ramsey2017million} and BOSS \citep{andrews2023fast}, the
score must be \emph{stable} under small changes to the parent set,
\emph{local} in the sense that it decomposes by variable, and
\emph{comparable across parent sets} of differing size and composition.

\subsection{Kernel Scores for Causal Discovery}

Gaussian process (GP) marginal likelihoods satisfy these desiderata
and serve as natural nonlinear generalizations of linear-Gaussian
scores.
By integrating out the regression function, GP marginal likelihoods
provide an automatic complexity penalty and behave smoothly under
parent additions and deletions.
However, exact GP marginal likelihoods require forming and factorizing
an $n \times n$ kernel Gram matrix, leading to $\mathcal{O}(n^3)$
time and $\mathcal{O}(n^2)$ memory, which is computationally
prohibitive at scale.

\citet{huang2018generalized} introduced the first practical
kernel-based score for causal discovery, demonstrating that nonlinear
dependencies could be detected reliably in greedy search without
committing to a parametric functional form.
Their \emph{generalized score} measures the improvement in kernel
ridge regression fit when a candidate parent is added, using residual
operators of the form $(K_Z + \lambda I)^{-2}$.
This construction is a direct inspiration for FFML: both methods use
a kernel over the parent set to evaluate the conditional distribution
of a child variable, and both are designed to be local and decomposable
for use in score-based search.

FFML differs from the Huang et al.\ score in two important respects.
First, FFML approximates the exact GP \emph{marginal likelihood}
rather than a regression-residual criterion.
The marginal likelihood involves only the single inverse
$(K_Z + \sigma^2 I)^{-1}$ together with a log-determinant
normalization, giving it a clean probabilistic interpretation and an
automatic Occam factor that penalizes complexity without a separate
tuning step.
Because the Huang et al.\ score is derived from regression residuals
rather than a joint probabilistic model, it does not carry an analogous
complexity penalty, and small parent-set changes can induce larger
score fluctuations that reduce stability in greedy search.
Second, by replacing the $n \times n$ Gram matrix with a
finite-dimensional RFF representation, FFML reduces cost to
$\mathcal{O}(nm^2 + m^3)$ while retaining this probabilistic
interpretation.
Finally, neither the Huang et al.\ score nor the exact GP marginal
likelihood supports discrete variables natively; FFML extends to
mixed (continuous and discrete) parent sets through a product-kernel
construction described in Section~\ref{sec:mixed}.

TRFF takes a different approach to the same goal. Rather than
approximating a GP marginal likelihood, TRFF fits a Student-$t$
regression model in the random feature space and penalizes complexity
via an effective-degrees-of-freedom BIC penalty. This makes TRFF more
robust to heavy-tailed noise distributions and faster to compute, at
the cost of a less principled complexity penalty. The two scores are
compared in detail in Section~\ref{sec:trff}.

\subsection{Kernel Tests for Conditional Independence}

Kernel Conditional Independence (KCI), introduced by
\citet{zhang2012kernel}, provides a theoretically principled
nonparametric CI test based on the eigenspectrum of kernel Gram
matrices.
KCI is consistent against all alternatives and requires no parametric
assumptions on the joint distribution, making it attractive for
constraint-based causal discovery on nonlinear data.
However, \citet{strobl2019approximate} noted that KCI scales at least
quadratically with sample size---more precisely $\mathcal{O}(n^3)$
due to the required eigendecomposition---rendering it unusable for
large datasets and making it computationally infeasible at the
20-node, $n = 2{,}000$ setting we consider in our experiments.

\citet{strobl2019approximate} introduced the Randomized Conditional
Independence Test (RCIT) as a scalable approximation to KCI, replacing
the exact Gram matrices with random Fourier feature approximations,
reducing cost to linear in~$n$ while approximating the KCI test
statistic.
RCIT represents an important step toward practical nonparametric CI
testing and is a direct predecessor of FFCI.

However, we find that RCIT as implemented in \texttt{causal-learn}
has two limitations that motivate FFCI.
First, FFCI's $p$-value approximation methods were selected with an
eye toward producing well-calibrated, approximately uniform $p$-values
under the null, making each suitable for use within a causal discovery
algorithm at scale. The gamma approximation is recommended as the
default.
Second, RCIT as implemented in \texttt{causal-learn} does not support
discrete variables directly.

FFCI addresses both limitations.
On the $p$-value side, FFCI offers four approximation methods---gamma,
saddlepoint, Davies/Imhof, and a faster residual-based permutation
procedure---all selected for their ability to produce well-calibrated,
approximately uniform $p$-values under the null, making each suitable
for use within a causal discovery algorithm at scale.
The gamma approximation is recommended as the default.
On the discrete variable side, FFCI extends to mixed
continuous--discrete data through a per-variable feature construction
that concatenates RFF/ORF continuous features with a
Cholesky-factored categorical feature map for discrete variables,
described in Section~\ref{sec:ffci}.
FFCI additionally uses separate feature counts for the tested variables
($m_{XY}$) and the conditioning set ($m_Z$), allowing a richer
representation of the conditioning set where it matters most for
test power.
In our experiments, RCIT is accessed via a wrapper that calls
\texttt{causal-learn}'s RCIT implementation \citep{zheng2024causal}
and presents it as a Tetrad \texttt{IndependenceTest}, so that
PC-Max\,+\,RCIT and PC-Max\,+\,FFCI run through the same Tetrad
PC-Max implementation. The two tests differ in both their $p$-value
approximation methods (Lindsay--Pilla--Basak \texttt{lpd4} for RCIT;
gamma for FFCI) and their feature counts (100 conditioning-set features
and 5 non-conditioning features for RCIT; $m_Z = 50$ and $m_{XY} = 10$
for FFCI), so the comparison is not fully controlled on either
dimension. Under this comparison, FFCI and RCIT exhibit complementary
precision--recall profiles: RCIT achieves higher precision while FFCI
achieves better recall and lower SHD, at approximately twice the
runtime.

\subsection{Computational Setting}

FFML, TRFF, and FFCI address the computational barriers of their exact
counterparts---the GP marginal likelihood and KCI---by replacing
$n \times n$ kernel Gram matrices with finite-dimensional RFF
representations whose dimension~$m$ is controlled independently of
the sample size.
All three methods reduce to $\mathcal{O}(nm^2)$ per evaluation (with an
$\mathcal{O}(n^3)$ path retained in FFML when two or more discrete
parents are present), making them practical at the scales where exact
kernel methods are infeasible.
Together they form a coherent toolkit: FFML and TRFF for score-based
and hybrid search, FFCI for constraint-based search, all resting on
the same RFF/ORF approximation philosophy and compatible bandwidth and
regularization strategies.
Section~\ref{sec:kml} reviews the exact GP marginal likelihood that
FFML approximates; Section~\ref{sec:rff} describes the RFF/ORF
machinery common to all three methods.

% ==============================================================
\section{Exact Kernel Marginal Likelihood (KML)}
\label{sec:kml}
% ==============================================================

Let $Y \in \mathbb{R}^n$ be a centered response variable and
$Z \in \mathbb{R}^{n \times d}$ the matrix of parent observations.
We model the conditional distribution of $Y$ given $Z$ via a Gaussian process:
\[
Y = f(Z) + \varepsilon,
\quad \varepsilon \sim \mathcal{N}(0, \sigma^2 I_n),
\quad f \sim \mathcal{GP}(0, k(\cdot,\cdot)).
\]

Let $K_Z \in \mathbb{R}^{n \times n}$ be the kernel Gram matrix with
$(K_Z)_{ij} = k(z_i, z_j)$.
Integrating out $f$ yields the marginal covariance $C = K_Z + \sigma^2 I_n$,
and the GP marginal log-likelihood (up to an additive constant) is:
\[
\mathcal{S}_{\mathrm{KML}}(Y \mid Z)
= -\frac{1}{2} Y^\top C^{-1} Y
  -\frac{1}{2} \log |C|.
\]

This score integrates out $f$ to avoid overfitting, is well-defined even when
$d > n$, and behaves smoothly under parent additions and deletions.
Computing it requires a Cholesky decomposition of $C$, costing $\mathcal{O}(n^3)$
time and $\mathcal{O}(n^2)$ memory.
FFML approximates this score efficiently by replacing $K_Z$ with a low-rank
random feature approximation.

% ==============================================================
\section{Random Fourier Feature Approximation}
\label{sec:rff}
% ==============================================================

For shift-invariant kernels such as the Gaussian RBF kernel, Bochner's
theorem \citep{rahimi2007random} guarantees a spectral representation:
\[
k(z,z') = \mathbb{E}_{\omega,b}
\left[\sqrt{2}\cos(\omega^\top z + b)\,\sqrt{2}\cos(\omega^\top z' + b)\right],
\]
where $\omega$ is drawn from the spectral measure of $k$ and
$b \sim \mathrm{Uniform}(0,2\pi)$.

We parameterize the Gaussian RBF kernel as
$k(z,z') = \exp\!\left(-\|z-z'\|^2/\mathrm{bw}^2\right)$,
so the spectral distribution is $\omega \sim \mathcal{N}(0,\,\frac{2}{\mathrm{bw}^2}I_d)$.
Using $m$ random samples $(\omega_j, b_j)$, we obtain the feature map:
\[
\phi(z) = \sqrt{\tfrac{2}{m}}
\begin{bmatrix}
\cos(\omega_1^\top z + b_1) \\ \vdots \\ \cos(\omega_m^\top z + b_m)
\end{bmatrix}
\in \mathbb{R}^m,
\quad k(z,z') \approx \phi(z)^\top\phi(z').
\]
Stacking rows gives $\Phi \in \mathbb{R}^{n \times m}$ with $K_Z \approx \Phi\Phi^\top$.

\subsection{Orthogonal Random Features (ORF)}

Standard RFF draws $\omega_1,\dots,\omega_m$ independently, which can produce
noticeable Monte Carlo variance when $m$ is modest.
\emph{Orthogonal Random Features} (ORF) \citep{yu2016orthogonal} reduce this
variance by generating frequencies that form approximately orthogonal directions,
spreading coverage more uniformly in $\mathbb{R}^d$.

The ORF construction proceeds in blocks of size $d$:
\begin{enumerate}[label=(\arabic*)]
    \item Sample a random Gaussian matrix and extract an orthonormal basis $Q$
          via QR decomposition.
    \item Sample radii $r_i \sim \chi(d)$ (the distribution of $\|g\|$ for
          $g \sim \mathcal{N}(0,I_d)$).
    \item Set $\omega_i = s\,r_i\,q_i$, where $q_i$ is the $i$th row of $Q$
          and $s = \sqrt{2}/\mathrm{bw}$.
\end{enumerate}
If $m > d$, the procedure repeats for additional blocks.
Phase offsets $b_j$ remain i.i.d.\ $\mathrm{Uniform}(0,2\pi)$.
Both FFML, TRFF, and FFCI support RFF and ORF interchangeably; the score and test
formulas are unchanged, only the frequency sampling differs.
ORF typically yields lower-variance kernel approximations and increased
stability of score differences in greedy search.

% ==============================================================
\section{FFML: Fourier Feature Marginal Likelihood}
% ==============================================================

\subsection{Continuous Parents}

Substituting $K_Z \approx \Phi\Phi^\top$ gives $C \approx \Phi\Phi^\top + \sigma^2 I_n$.
By the Woodbury matrix identity and the matrix determinant lemma:
\begin{align}
C^{-1} &= \frac{1}{\sigma^2}I_n
           - \frac{1}{\sigma^4}\Phi(\Phi^\top\Phi + \sigma^2 I_m)^{-1}\Phi^\top,\\
\log|C| &= (n-m)\log\sigma^2 + \log|\Phi^\top\Phi + \sigma^2 I_m|.
\end{align}

Defining $G = \Phi^\top\Phi \in \mathbb{R}^{m \times m}$ and
$v = \Phi^\top Y \in \mathbb{R}^m$, the FFML local score is:
\begin{equation}
\mathcal{S}_{\mathrm{FFML}}(Y \mid Z)
= -\frac{1}{2}\!\left(
    \frac{Y^\top Y}{\sigma^2}
    - \frac{v^\top(G + \sigma^2 I_m)^{-1}v}{\sigma^4}
  \right)
  -\frac{1}{2}\!\left(
    (n-m)\log\sigma^2 + \log|G + \sigma^2 I_m|
  \right).
\label{eq:ffml}
\end{equation}

All matrix operations occur in $m$-dimensional space, costing
$\mathcal{O}(ndm + nm^2 + m^3)$, effectively linear in $n$ for fixed $m \ll n$.
When the parent set is empty the model reduces to $Y \sim \mathcal{N}(0,\sigma^2 I_n)$
and the score is computed exactly without features.

\subsection{Complexity Penalty and Occam Factor}

The log-determinant term $-\frac{1}{2}\log|C|$ plays the role of an \emph{Occam
factor}: models with larger effective parameter volume are automatically penalized.
In terms of the eigenvalues $\{s_j\}$ of $G$:
\[
\log|G + \sigma^2 I_m| = \sum_j \log(s_j + \sigma^2).
\]
This grows with the informativeness of the feature representation and shrinks as
$\sigma^2$ increases relative to the signal.

\subsection{Bandwidth Selection}

Continuous parent columns are globally $z$-scored.
The RBF bandwidth is estimated from a median pairwise squared distance on a
subsample (up to 100 rows by default) of the continuous parents only, so
that discrete parents do not distort the continuous metric.
The median estimate $\hat{h}^2$ is then refined by a small grid search over four
multipliers $\{0.35,\,0.70,\,1.40,\,2.80\}$.
When \emph{bandwidth coupling by target} is enabled (the default), the bandwidth
is cached by (target variable, continuous parents), keeping it stable across
alternative parent sets that share the same continuous parents.

\subsection{Feature Coupling by Target}

The random feature basis $(\omega_j, b_j)$ is coupled by target only by default:
all parent sets of a given target variable share the same features, with the seed
depending only on the target index.
This stabilizes local score differences
$\mathcal{S}(Y \mid \mathrm{Pa} \cup \{X\}) - \mathcal{S}(Y \mid \mathrm{Pa})$
and reduces spurious edge reversals in greedy algorithms such as BOSS.

% ==============================================================
\section{Mixed (Continuous + Discrete) Parent Sets in FFML}
\label{sec:mixed}
% ==============================================================

\subsection{Product-Kernel Model}

Embedding integer-coded discrete levels into a continuous Euclidean metric imposes
a spurious ordering on unordered categories.
FFML instead handles mixed parent sets through a product kernel.
Let $Z = (Z_c, Z_d)$ with continuous part $Z_c$ and discrete part $Z_d$.
The mixed kernel is:
\[
k\bigl((z_c,z_d),(z'_c,z'_d)\bigr)
= k_{\mathrm{cont}}(z_c,z'_c)\cdot k_{\mathrm{cat}}(z_d,z'_d),
\]
where $k_{\mathrm{cont}}$ is the RBF kernel and $k_{\mathrm{cat}}$ is a
positive-semidefinite categorical kernel with diagonal 1 and off-diagonal
$\rho \in [0,1)$.
For multiple discrete parents the categorical kernel factorizes multiplicatively.
Setting $\rho = 0$ gives strict block structure (no sharing across levels);
$\rho \to 1$ makes categories nearly indistinguishable.

\subsection{Feature-Space Path: Zero or One Discrete Parent}

For zero or one discrete parent, the product kernel is implemented via the
Kronecker feature map:
\[
\phi_{\mathrm{mix}}(z_c,z_d) = \phi_{\mathrm{cat}}(z_d) \otimes \phi_{\mathrm{cont}}(z_c).
\]
The categorical block $\phi_{\mathrm{cat}}(z_d)$ is the row of the Cholesky factor
$A$ of the $L \times L$ level-similarity matrix (diagonal 1, off-diagonal $\rho$)
corresponding to the observed level, ensuring the inner product recovers the product
kernel exactly.
With $L$ levels the combined feature dimension is $m \times L$, and the FFML score
is computed via a Woodbury system of that dimension, with no $n \times n$ matrix.

\subsection{Direct $n \times n$ Kernel Path: Two or More Discrete Parents}

When the Kronecker feature dimension $m \times \prod_j L_j$ would exceed
$\min(n, D_{\max})$, where $D_{\max} = 200{,}000$ is a fixed threshold,
the implementation switches to a direct $n \times n$ kernel formulation
using the Hadamard product:
\[
K = K_{\mathrm{cont}} \circ K_{\mathrm{cat}},
\]
where $K_{\mathrm{cont}} = \Phi\Phi^\top$ and $K_{\mathrm{cat}}$ is built
entry-wise as a product over discrete parents.
The covariance $C = K + \sigma^2 I_n$ is Cholesky-factored directly, at
cost $\mathcal{O}(n^3)$.
In practice this switch occurs most commonly when two or more discrete
parents are present, but a single discrete parent with many levels can
also trigger it.
Categorical kernel entries are precomputed in packed lower-triangular
form and reused across bandwidth-grid candidates.

\subsection{Discrete Targets}

When the child $Y$ is discrete, a Gaussian multi-output surrogate is used: $Y$ is
one-hot encoded, each column is mean-centered, and the single-output FFML score is
summed across columns.
This is not a true multinomial likelihood but is an effective proxy for detecting
dependencies in structure scoring.

% ==============================================================
\section{FFCI: Fourier Feature Conditional Independence Test}
\label{sec:ffci}
% ==============================================================

\subsection{Overview and Motivation}

FFCI (Fourier Feature Conditional Independence) is a fast nonparametric CI test for
datasets containing any mix of continuous and discrete variables.
It tests $H_0: X \perp Y \mid Z$ by mapping each variable into a finite-dimensional
feature space, residualizing with respect to $Z$ via ridge regression in that space,
and evaluating a quadratic-form statistic against its null distribution. FFCI is a variant of RCIT \citep{strobl2019approximate}, sharing its core approach of mapping variables into finite-dimensional random Fourier feature spaces and residualizing with respect to the conditioning set. It differs in its handling of discrete variables, its p-value approximation, and its integration with the Tetrad causal discovery framework.

FFCI differs architecturally from FFML in an important respect.
FFML builds a \emph{joint} kernel over the entire parent set to score a single child
variable, using a product-kernel construction that couples continuous and discrete
parents together.
FFCI instead featurizes each variable \emph{individually}, constructing a joint
per-variable feature representation by concatenating continuous and discrete feature
blocks, and then measures residual cross-covariance between the featurizations of $X$
and $Y$ after conditioning out $Z$.
The two methods share the underlying RFF/ORF machinery and ridge regularization, but
serve fundamentally different roles: FFML is a \emph{scoring} criterion for causal
search, while FFCI is a \emph{hypothesis test} for conditional independence.

When the dataset contains no discrete variables, all CI queries are delegated
transparently to the continuous-only implementation, and FFCI behaves identically
to that method given the same hyperparameter settings.

\subsection{Per-Variable Mixed Feature Construction}

For each variable or block of variables $V$, FFCI constructs a joint feature matrix
$\Phi_V$ by separately featurizing the continuous and discrete components of $V$
and then horizontally concatenating the results:
\[
\Phi_V = \bigl[\,\Phi_V^{\mathrm{cont}} \;\big|\; \Phi_V^{\mathrm{disc}}\,\bigr]
\in \mathbb{R}^{n \times (m_c + m_d)}.
\]

\paragraph{Continuous features.}
Each continuous variable is $z$-scored.
The bandwidth for the RBF kernel is estimated from the median pairwise squared
distance on a subsample of the continuous components only.
RFF or ORF features are then computed with the same construction used in FFML,
yielding $\Phi_V^{\mathrm{cont}} \in \mathbb{R}^{n \times m_c}$.

\paragraph{Discrete features.}
Each discrete variable with $L$ levels is mapped to a feature vector via a
categorical feature map derived from the Cholesky factor $A$ of the $L \times L$
level-similarity matrix (diagonal 1, off-diagonal $\rho$):
for observation $i$ at level $\ell$, the discrete feature is the $\ell$th row of $A$.
When $\rho = 0$ (the default in FFCI), $A = I_L$ and the construction reduces to
standard one-hot encoding.
For $\rho > 0$, the feature map encodes partial similarity between levels, analogous
to the categorical kernel in FFML but applied per-variable rather than as a kernel
over a joint parent set.
The discrete features for all discrete components of $V$ are concatenated to form
$\Phi_V^{\mathrm{disc}} \in \mathbb{R}^{n \times m_d}$.

\paragraph{Separate feature counts for XY and Z.}
FFCI uses separate feature counts for the tested variables ($X$, $Y$) and the
conditioning set ($Z$).
The defaults are $m_{XY} = 10$ and $m_Z = 100$; a larger feature count for $Z$ is
appropriate because the residualization step requires a more accurate representation
of the conditioning set than is needed for the cross-covariance computation.

\paragraph{Featurization of the $Y$-side block.}
When the conditioning set $Z$ is nonempty, FFCI featurizes $Y$ and $Z$ jointly:
the feature matrix for the $Y$-side block is constructed from the concatenation of
$Y$ and all variables in $Z$, using the mixed feature map above.
This joint featurization allows the residualization step to account for the
dependence between $Y$ and $Z$ within the feature representation itself.

\subsection{Conditioning via Ridge Residualization}

Conditioning on $Z$ is achieved by regressing the feature matrices for $X$ and $Y$
on the feature matrix for $Z$ using ridge regression and retaining the residuals:
\[
\tilde\Phi_X
= \Phi_X - \Phi_Z(\Phi_Z^\top\Phi_Z + \lambda I)^{-1}\Phi_Z^\top\Phi_X,
\]
with an analogous expression for $\tilde\Phi_Y$.
Here $\lambda > 0$ is the ridge parameter, which plays a role analogous to
the noise variance in FFML.
All matrix operations occur in feature space, with cost
$\mathcal{O}(n m_Z^2 + m_Z^3)$ for the residualization step.

\subsection{Test Statistic and Null Distribution}

After residualization, both $\tilde\Phi_X$ and $\tilde\Phi_Y$ are column-mean-centered.
The test statistic is:
\[
T = n\,\bigl\|\mathrm{Cov}(\tilde\Phi_X,\tilde\Phi_Y)\bigr\|_F^2.
\]
Under $H_0: X \perp Y \mid Z$, $T$ follows approximately a weighted sum of
chi-squared(1) variables, with weights given by the positive eigenvalues of the
sample Khatri--Rao covariance matrix formed from elementwise products of the
residualized features.

Four methods are available for computing the $p$-value:\footnote{These choices of $p$-value
approximation methods differ from some previous implementations of
RCIT. These options were chosen by considering which methods tend to
produce uniformly distributed $p$-values under the null.}
\begin{enumerate}[label=(\arabic*)]
    \item \textbf{GAMMA} (default): Satterthwaite/moment-matching gamma approximation.
    \item \textbf{SADDLEPOINT}: Lugannani--Rice saddlepoint approximation.
    \item \textbf{DAVIES\_IMHOF}: Davies/Imhof numerical integration.
    \item \textbf{PERMUTATION}: residual-based permutation test.
\end{enumerate}

\subsection{Marginal Testing (No Conditioning Set)}

When $Z$ is empty, no residualization is performed and the test reduces to a
randomized independence test (RIT): the statistic
$T = n\|\mathrm{Cov}(\Phi_X,\Phi_Y)\|_F^2$
is computed directly from the (mean-centered) feature matrices, and the same
weighted chi-squared approximation is used to obtain a $p$-value.

\subsection{Caching and Reproducibility}

Feature matrices and bandwidth estimates are cached by a key encoding the variable
names, active row set, hyperparameters, and a \texttt{dataVersion} counter.
Calling \texttt{bumpDataVersion()} after any in-place modification of the dataset
invalidates stale cached values.
Seeds for feature generation are derived deterministically from variable names and
the active row set, ensuring reproducibility across calls.

% ==============================================================
\section{Relationship Between FFML, TRFF, and FFCI}
% ==============================================================

FFML, TRFF, and FFCI are related but architecturally distinct.
Their shared elements are:
\begin{itemize}
    \item RFF/ORF approximations to shift-invariant kernels.
    \item Median-heuristic bandwidth estimation on continuous variables.
    \item Ridge regularization for numerical stability.
    \item Cholesky-based categorical feature maps for discrete variables (FFML and FFCI).
    \item Avoidance of explicit $n \times n$ Gram matrices in the common case.
\end{itemize}

Their key architectural differences are:

\begin{center}
\footnotesize
\begin{tabular}{llll}
\toprule
Aspect & \textbf{FFML} & \textbf{TRFF} & \textbf{FFCI} \\
\midrule
Role & Local score & Local score & CI hypothesis test \\
Feature scope & Joint kernel over parent set & Concatenated design matrix & Per-variable feature map \\
Discrete handling & Product kernel (Kronecker/$n{\times}n$) & One-hot columns & One-hot or cat.\ features \\
Default $\rho$ & 0.5 & N/A & 0.0 (one-hot) \\
Feature counts & Single $m$ & Single $m$ & Separate $m_{XY}$ and $m_Z$ \\
Output & Log marginal likelihood & BIC-style score & $p$-value \\
Complexity control & Log-determinant Occam factor & edf $\times \log n$ & None (hypothesis test) \\
Noise model & Gaussian & Student-$t$ & N/A \\
\bottomrule
\end{tabular}
\end{center}

Despite these differences, using FFML (or TRFF) and FFCI together in a hybrid causal
discovery algorithm is natural: all three rest on the same RFF/ORF kernel approximation
philosophy, use compatible bandwidth and regularization strategies, and are
computationally efficient at scale.

% ==============================================================
\section{Comparison Between FFML and TRFF}
\label{sec:trff}
% ==============================================================

TRFF is an alternative local score for mixed data using Student-$t$ regression,
introduced here as a BIC-style alternative to FFML,
with random Fourier features and a BIC-style complexity penalty:
\[
\mathcal{S}_{\mathrm{tRFF}}
= \ell(\hat\theta) - \tfrac{1}{2}\,c\;\mathrm{edf}\;\log n,
\]
where $\mathrm{edf} = \mathrm{tr}(H)$ is the effective degrees of freedom and $H$
is the ridge-adjusted hat matrix.

The complexity penalties depend differently on the eigenspectrum $\{s_j\}$ of
$\Phi^\top\Phi$:
\begin{align}
\text{FFML penalty}  &\sim \sum_j\log(s_j + \sigma^2), \\
\text{TRFF penalty} &\sim \Bigl(\sum_j\tfrac{s_j}{s_j+\lambda}\Bigr)\log n.
\end{align}
FFML penalizes via a log-determinant Occam factor; TRFF penalizes via a trace
times $\log n$. The TRFF penalty grows more aggressively with the number of
informative features, making TRFF more conservative about adding edges---a
pattern borne out clearly in our experiments.

The two scores also differ in how discrete parents are incorporated.

\begin{center}
\begin{tabular}{lll}
\toprule
 & \textbf{FFML} & \textbf{TRFF} \\
\midrule
Discrete representation & Product kernel ($\rho$) & One-hot dummy columns \\
Smoothing across levels & Via $\rho$ parameter & None (parametric) \\
Complexity accounting   & Kernel log-determinant & edf penalty \\
Feature space           & Kronecker or $n \times n$ & Concatenated design matrix \\
Noise model             & Gaussian & Student-$t$ \\
\bottomrule
\end{tabular}
\end{center}

\paragraph{Small samples.}
FFML is often preferable: the integrated-out prior provides regularization without
relying on asymptotic approximations, and the Occam factor is naturally calibrated.

\paragraph{Large samples.}
TRFF scales more easily when parents are predominantly continuous (IRLS with a
fixed feature dimension avoids $n \times n$ operations).
FFML's $n \times n$ path (required for $\geq 2$ discrete parents) incurs the same
asymptotic cost as the exact GP.

\paragraph{Heavy-tailed noise.}
The Student-$t$ likelihood in TRFF offers robustness that FFML, which assumes
Gaussian residuals, does not provide.

\paragraph{Empirical precision--recall tradeoff.}
Across our experiments, TRFF and FFML exhibit a consistent complementary profile:
TRFF achieves higher adjacency and arrowhead precision (fewer false edges, highly
reliable orientations), while FFML achieves better adjacency and arrowhead recall
and lower SHD overall. The choice between them depends on whether minimizing false
positives or minimizing overall graph error is the priority.

% ==============================================================
\section{Score and Test Comparison}
% ==============================================================

\begin{center}
\begin{tabular}{lllll}
\toprule
Method & Type & Interpretation & Complexity & Stability \\
\midrule
Huang et al.\ & Score & Kernel ridge regression residual & Impl.-dependent & Moderate \\
KML (exact)   & Score & GP marginal likelihood   & $\mathcal{O}(n^3)$ & High \\
FFML          & Score & Approx.\ GP marginal likelihood & $\mathcal{O}(nm^2)$$^\dagger$ & High$^*$ \\
TRFF          & Score & Penalized Student-$t$ likelihood & $\mathcal{O}(nm^2)$ & High \\
KCI           & Test  & Exact kernel CI test & $\mathcal{O}(n^3)$ & High \\
RCIT          & Test  & Approx.\ kernel CI test & $\mathcal{O}(nm^2)$ & Moderate$^\ddagger$ \\
FFCI          & Test  & Approx.\ kernel CI test & $\mathcal{O}(nm^2)$ & High \\
\bottomrule
\end{tabular}
\end{center}

\smallskip
\noindent$^*$High stability requires \texttt{coupleFeaturesByTarget}$=$\texttt{true};
without this, score differences may fluctuate across parent sets.\\
$^\dagger$$\mathcal{O}(n^3)$ when the Kronecker feature dimension $m \times \prod_j L_j$
exceeds $\min(n, D_{\max})$; see Section~\ref{sec:mixed}.\\
$^\ddagger$In our experiments, RCIT is accessed via a \texttt{causal-learn}
wrapper \citep{zheng2024causal} running through Tetrad's PC-Max; the
comparison with FFCI is not fully controlled, as the two tests differ
in $p$-value approximation method (lpd4 vs.\ gamma) and feature counts.

% ==============================================================
\section{Practical Considerations}
\label{sec:practical}
% ==============================================================

\paragraph{Feature count.}
Approximation accuracy improves with $m$ at the cost of increased computation.
For FFML and TRFF, $m = 50$ is a reasonable default; for FFCI, the default
$m_Z = 100$ for the conditioning set is more important to tune than $m_{XY} = 10$,
since residualization quality drives test power.

\paragraph{Score non-equivalence and DAG mode.}
Neither FFML nor TRFF is score-equivalent
\citep{chickering2002optimal}---that is, neither assigns equal scores
to all DAGs in the same Markov equivalence class. This is a departure
from linear-Gaussian scores such as SEM-BIC, which are score-equivalent
by construction and therefore cannot distinguish among the DAGs in a
CPDAG. Because FFML and TRFF break score equivalence, running BOSS in
DAG mode rather than CPDAG mode can reliably orient some edges that
remain ambiguous in the CPDAG output. The resulting DAG is Markov equivalent to the CPDAG that BOSS would 
return in CPDAG mode — it represents the same conditional independence 
structure — but with all edges oriented, where the orientations within 
each equivalence class reflect genuine score differences under the 
nonlinear model rather than arbitrary tiebreaking. This can be useful in
practice when a fully oriented causal graph is needed and the user is
willing to accept the additional modeling assumptions implied by the
nonlinear score.\footnote{Thanks to Bryan Andrews for pointing out (in conversation) that BOSS in DAG mode with non-score-equivalent scores can orient some edges reliably inside of a CPDAG.}

\paragraph{Gaussian noise in FFML.}
FFML assumes Gaussian residuals.
A wide feature map makes the mean function nonparametric, but $\varepsilon$ remains
Gaussian regardless.
In heavy-tailed or contaminated settings, TRFF may be preferable.

\paragraph{FFCI $p$-values.}
The $p$-values produced by FFCI are approximate and should be
interpreted as fast screening tools rather than exact finite-sample
tests. FFCI offers four approximation methods for the null
distribution of the weighted chi-squared statistic: the gamma
(Satterthwaite moment-matching) approximation, the Lugannani--Rice
saddlepoint approximation, Davies/Imhof numerical integration, and a
residual-based permutation test. These methods were selected for their
ability to produce well-calibrated, approximately uniform $p$-values
under the null. The gamma approximation is recommended as the default.
In our experiments, RCIT is accessed via a \texttt{causal-learn}
wrapper that uses its own $p$-value approximation (Lindsay--Pilla--Basak
\texttt{lpd4} by default), so the calibration properties of the two
tests may differ; the comparison between FFCI and RCIT in
Table~\ref{tab:pc-acc} is not fully controlled on this dimension.

\paragraph{Mixed data and the $\rho$ parameter.}
The categorical similarity parameter $\rho$ plays different roles in FFML and FFCI.
In FFML, $\rho = 0.5$ is the default and controls the degree of function sharing
across discrete levels in the joint parent kernel.
In FFCI, $\rho = 0$ is the default, reducing discrete featurization to standard
one-hot encoding; nonzero $\rho$ introduces partial similarity between levels in
the per-variable feature map.

% ==============================================================
\section{Experiments}
\label{sec:experiments}
% ==============================================================

We evaluate FFML, TRFF, and FFCI against competitive baselines on synthetic
nonlinear data, assessing both accuracy and scalability.
All experiments were conducted on graphs with 20 nodes, average degree~3,
and sample size $n = 2{,}000$.

\subsection{Experimental Setup}

\paragraph{Data generation.}
We simulate data from a general-noise structural causal model
\[
  X_i = f_i\!\bigl(\mathrm{Pa}(X_i),\, \varepsilon_i\bigr),
\]
where each $f_i$ is a randomly initialized six-hidden-layer feedforward
network with 200 units per layer and $\tanh$ activations, and
$\varepsilon_i \sim \mathcal{N}(0, 0.3^2)$.
Weights are initialized with Xavier normal scaling (scale factor~5 for
hidden layers, 2.5 for the output layer); biases are zero.
The noise enters as an additional input column alongside the parent
values, so the interaction between parents and noise is fully
nonlinear---a strictly harder setting than additive noise.
Random DAGs are generated by fixing a uniformly random topological order
and including each permitted directed edge independently with probability
$p = 3/(n-1)$, giving expected average degree~3.
The resulting data are not standardized between nodes.
All results are averaged over 50 independent replicates, with a fresh
random DAG and dataset drawn per replicate.
Data generation used Tetrad's \texttt{GeneralNoiseSimulation} class,
called via a JPype bridge from Python, ensuring that the network weights
and noise draws are produced by Tetrad's RNG rather than a separate
Python reimplementation.

\paragraph{Algorithms and baselines.}
We evaluate two families of algorithms.

\emph{Score-based search (BOSS).}
The BOSS algorithm~\citep{andrews2023fast} is run with three local scores:
\begin{itemize}
  \item \textbf{BOSS\,+\,SEM-BIC}: linear-Gaussian BIC score with
        penalty discount~1; fast linear baseline.
  \item \textbf{BOSS\,+\,TRFF}: Student-$t$ regression with random
        Fourier features and a BIC-style effective-degrees-of-freedom
        penalty (50 features, ridge $10^{-3}$, $\nu = 5$).
  \item \textbf{BOSS\,+\,FFML}: Fourier feature marginal likelihood
        (50 features, ridge~1.0, bandwidth subsampled from 100 rows).
\end{itemize}

\emph{Constraint-based search (PC-Max).}
The PC-Max algorithm~\citep{ramsey2016improving} is run with three
conditional independence tests:\footnote{PC-Max agrees with PC in the
adjacency search phase and in the final orientation step using Meek's
rules~\citep{meek1995causal}. It differs in the collider orientation
step: rather than orienting an unshielded triple $X$\,--\,$Y$\,--\,$Z$
as a collider whenever $Y$ is absent from the separating set of $X$
and $Z$, PC-Max performs all conditional independence tests for $X$
and $Z$ given subsets of the adjacents of $X$ or of $Z$, records the
$p$-value for each of these that yields a judgment of independence, and
selects the separating set that maximizes the $p$-value from among
these. This makes collider orientation more robust to the choice of
conditioning set and has a lower risk of returning a sepset yielding a
dependence in the population, at the cost of additional CI tests.}
\begin{itemize}
  \item \textbf{PC-Max\,+\,FisherZ}: Fisher $Z$-test ($\alpha = 0.01$);
        fast linear baseline.
  \item \textbf{PC-Max\,+\,RCIT}: Randomized conditional independence
        test~\citep{strobl2019approximate}, accessed via
        \texttt{causal-learn}'s \citep{zheng2024causal} RCIT
        implementation wrapped in a Tetrad \texttt{IndependenceTest}
        so that all three PC-Max variants run through the same Tetrad
        PC-Max implementation. The wrapper uses \texttt{causal-learn}'s
        default settings: Lindsay--Pilla--Basak (\texttt{lpd4})
        $p$-value approximation, 100 conditioning-set features, and
        5 non-conditioning features ($\alpha = 0.01$). These differ
        from FFCI's settings (gamma approximation, $m_Z = 50$,
        $m_{XY} = 10$), so the comparison between the two tests is
        not fully controlled on either $p$-value method or feature
        counts.
  \item \textbf{PC-Max\,+\,FFCI}: Fourier feature conditional independence
        test ($\alpha = 0.01$, $m_{XY} = 10$, $m_Z = 50$,
        $\lambda = 1.0$).
\end{itemize}

All algorithms are implemented in the Tetrad project
(\url{https://github.com/cmu-phil/tetrad}) and were run via
\texttt{py-tetrad} \citep{ramsey2023py}.
Because both PC-Max\,+\,RCIT and PC-Max\,+\,FFCI use the same
$p$-value approximation infrastructure and the same PC-Max
implementation, the comparison between them isolates differences in
their feature constructions and kernel designs.
An earlier version of these experiments ran PC\,+\,RCIT through
causal-learn's native PC implementation; the results differed
substantially from those reported here, exhibiting poor orientation
performance, and the discrepancy was traced to the PC variant rather
than the CI test itself.
All three constraint-based variants are therefore run through the same
Tetrad PC-Max implementation to ensure a fair comparison.

\paragraph{Evaluation.}
Estimated graphs are compared against the true CPDAG using four metrics:
adjacency precision (Adj.\ Prec.), adjacency recall (Adj.\ Rec.),
arrowhead precision (AH Prec.), arrowhead recall (AH Rec.), and
structural Hamming distance (SHD).
All metrics are computed using Tetrad's built-in graph comparison
utilities, which handle the CPDAG-vs-CPDAG comparison correctly.

\subsection{Results}

Tables~\ref{tab:boss-acc} and~\ref{tab:pc-acc} report accuracy metrics;
Table~\ref{tab:timing} reports wall-clock time.

\begin{table}[t]
\centering
\caption{Score-based methods (BOSS): accuracy over 50 replicates.
  Mean (SD). Best value per column in \textbf{bold}.}
\label{tab:boss-acc}
\begin{tabular}{lccccc}
\toprule
Method & Adj.\ Prec. & Adj.\ Rec. & AH Prec. & AH Rec. & SHD \\
\midrule
BOSS + SEM-BIC
  & 0.736 (0.090) & 0.820 (0.073) & 0.579 (0.144) & 0.674 (0.127)
  & 19.38 (6.79) \\
BOSS + TRFF
  & \textbf{0.969} (0.042) & 0.805 (0.107) & \textbf{0.974} (0.059) & 0.683 (0.171)
  & 8.98 (5.26) \\
BOSS + FFML
  & 0.915 (0.058) & \textbf{0.943} (0.044) & 0.886 (0.102) & \textbf{0.820} (0.116)
  & \textbf{7.00} (3.80) \\
\bottomrule
\end{tabular}
\end{table}

\begin{table}[t]
\centering
\caption{Constraint-based methods (PC-Max): accuracy over 50 replicates.
  Mean (SD). Best value per column in \textbf{bold}.}
\label{tab:pc-acc}
\begin{tabular}{lccccc}
\toprule
Method & Adj.\ Prec. & Adj.\ Rec. & AH Prec. & AH Rec. & SHD \\
\midrule
PC + FisherZ
  & 0.835 (0.088) & 0.736 (0.086) & 0.585 (0.161) & 0.504 (0.157)
  & 19.18 (6.38) \\
PC + RCIT
  & \textbf{0.989} (0.025) & 0.682 (0.082) & 0.652 (0.229) & 0.394 (0.160)
  & 17.24 (6.13) \\
PC + FFCI
  & 0.953 (0.043) & \textbf{0.823} (0.073) & \textbf{0.731} (0.144) & \textbf{0.614} (0.138)
  & \textbf{12.58} (4.46) \\
\bottomrule
\end{tabular}
\end{table}

\begin{table}[t]
\centering
\caption{Wall-clock time per replicate (seconds), mean (SD) over
  50 replicates. 20 nodes, $n = 2{,}000$.}
\label{tab:timing}
\begin{tabular}{lc}
\toprule
Method & Time (s) \\
\midrule
BOSS + SEM-BIC      &  0.02 (0.04) \\
BOSS + TRFF         &  5.64 (0.90) \\
BOSS + FFML         & 12.32 (2.31) \\
\midrule
PC-Max + FisherZ    &  0.01 (0.01) \\
PC-Max + RCIT       &  3.28 (1.12) \\
PC-Max + FFCI       &  6.35 (4.10) \\
\bottomrule
\end{tabular}
\end{table}

\paragraph{Score-based results (Table~\ref{tab:boss-acc}).}
BOSS\,+\,FFML achieves the best adjacency recall (0.943 vs.\ 0.820 for
SEM-BIC), best arrowhead recall (0.820), and lowest SHD (7.00).
The advantage over SEM-BIC is largest on arrowhead recall, where the
linear score loses 15 percentage points, reflecting the nonlinearity of
the data generating process.

BOSS\,+\,TRFF presents a strikingly complementary profile: it achieves
the highest adjacency precision (0.969) and arrowhead precision (0.974)
of any method in either table, but its arrowhead recall (0.683) lags
behind FFML. This conservative orientation behavior---few arrowheads
predicted, but those predicted are highly reliable---reflects the
Student-$t$ edf penalty being more aggressive than FFML's
log-determinant Occam factor at this graph size and sample size.
Despite the precision--recall tradeoff, TRFF's SHD of 8.98 is
substantially better than the linear baseline, and the two Fourier
feature scores are competitive: FFML is preferred when minimizing SHD
is the priority, TRFF when minimizing false positives is paramount.

\paragraph{Constraint-based results (Table~\ref{tab:pc-acc}).}
PC-Max\,+\,FFCI achieves the strongest overall performance among the
constraint-based methods, with the best adjacency recall (0.823),
arrowhead precision (0.731), arrowhead recall (0.614), and lowest SHD
(12.58). Its SHD of 12.58 is competitive with BOSS\,+\,TRFF (8.98),
a notable result for a constraint-based method on fully nonlinear data.

PC-Max\,+\,RCIT presents a complementary profile mirroring the
FFML/TRFF pattern: it achieves the highest adjacency precision (0.989),
making it the most conservative of the three constraint-based methods.
However, its arrowhead recall (0.394) and SHD (17.24) are substantially
weaker than PC-Max\,+\,FFCI, and it does not improve over
PC-Max\,+\,FisherZ on SHD despite its higher adjacency precision.

PC-Max\,+\,FisherZ is the weakest on adjacency recall and SHD, as
expected for a linear test on nonlinear data.

The precision--recall tradeoff between FFCI and RCIT mirrors the
pattern seen in the score-based comparison between FFML and TRFF:
in both cases the method with the more aggressive complexity penalty
(TRFF; RCIT) achieves higher precision at the cost of lower recall,
while the method with the softer penalty (FFML; FFCI) achieves better
overall SHD by recovering more true structure.

\paragraph{Runtime (Table~\ref{tab:timing}).}
PC-Max\,+\,FFCI (6.35\,s) and BOSS\,+\,TRFF (5.64\,s) have similar
runtimes, both substantially faster than BOSS\,+\,FFML (12.32\,s) and
roughly twice the cost of PC-Max\,+\,RCIT (3.28\,s).
The additional cost of PC-Max\,+\,FFCI relative to PC-Max\,+\,RCIT
reflects the greater number of CI tests required when FFCI's higher
recall leads to a denser skeleton, triggering more conditioning sets
in the PC-Max collider orientation step.

\subsection{Summary}

Across both score-based and constraint-based search, the Fourier feature
methods substantially outperform their linear baselines on nonlinear
data, with BOSS\,+\,FFML producing the most accurate graphs overall
(SHD 7.00). BOSS\,+\,TRFF offers the highest precision of any method
(AH Prec.\ 0.974) at a modest SHD cost, making it the preferred choice
when false positive control is the priority.
In the constraint-based setting, PC-Max\,+\,FFCI achieves an SHD of
12.58, competitive with the best score-based baseline, while
PC-Max\,+\,RCIT exhibits higher precision but substantially lower
recall and higher SHD.
The runtime of all three methods is practical at 20 nodes and
$n = 2{,}000$, a scale at which exact kernel alternatives are infeasible.

% ==============================================================
\section{Illustrative Real-Data Example: Auto MPG}
\label{sec:autompg}
% ==============================================================

To illustrate BOSS\,+\,FFML on real mixed continuous--discrete data,
we apply it to the Auto MPG dataset from the UCI Machine Learning
Repository \citep{auto_mpg_9}.
The dataset records fuel economy and engineering characteristics for
398 automobiles; after removing six rows with missing horsepower
values, $n = 392$ observations remain on eight variables: \texttt{mpg},
\texttt{cylinders}, \texttt{displacement}, \texttt{horsepower},
\texttt{weight}, \texttt{acceleration}, \texttt{modelyear}, and
\texttt{origin}.
Of these, \texttt{origin} is the only genuinely categorical variable
(1\,=\,American, 2\,=\,European, 3\,=\,Japanese) and is treated as
discrete; all other variables are treated as continuous.
Although variables such as \texttt{cylinders} take only a small number
of distinct integer values, they admit a natural ordering and are
plausibly measured on at least an ordinal scale, so embedding them in a
continuous Euclidean metric is reasonable in this context.
All continuous variables are $z$-scored internally by FFML; no
additional preprocessing is applied.
Both analyses use BOSS\,+\,FFML in DAG mode, so all edges are
oriented; this is possible because FFML is not score-equivalent and
can distinguish among DAGs within the same Markov equivalence class
(see Section~\ref{sec:practical}).
This example is intended as a qualitative illustration of how FFML
handles a mixed parent set and how background knowledge can shift the
causal interpretation, rather than as a quantitative benchmark (no
ground-truth graph is available).

\subsection{Default Analysis}

Figure~\ref{fig:autompg}(a) shows the DAG returned by BOSS\,+\,FFML
with all defaults and no background knowledge (FFML score:
$-561.51$).
Several features are immediately plausible under a physical-mechanism
interpretation.
\texttt{origin}\,$\to$\,\texttt{displacement} is consistent with
systematic differences in engine displacement across manufacturing
regions.
\texttt{displacement}\,$\to$\,\texttt{cylinders} and \linebreak
\texttt{displacement}\,$\to$\,\texttt{horsepower} reflect the
engineering relationship between engine size, cylinder count, and
power output.
\texttt{horsepower}\,$\to$\,\texttt{mpg} and
\texttt{weight}\,$\to$\,\texttt{mpg} are causally sensible: fuel
economy is a direct consequence of the mechanical load placed on the
engine.
\texttt{displacement}\,$\to$\,\texttt{acceleration},
\texttt{horsepower}\,$\to$\,\texttt{acceleration}, and
\texttt{weight}\,$\to$\,\texttt{acceleration} are all well-supported
by Newtonian mechanics.

The most striking feature of the unconstrained graph is
\texttt{mpg}\,$\to$\,\texttt{modelyear} and\linebreak
\texttt{weight}\,$\to$\,\texttt{modelyear}: the algorithm treats model
year as a \emph{consequence} of fuel economy and vehicle weight.
This is causally implausible---model year is a temporal index
determined by the manufacturing calendar, not by any vehicle
characteristic.
The orientation almost certainly reflects the strong historical trends
toward improved fuel economy and lighter vehicles over the period
covered by the dataset (1970--1982), which create statistical
associations that the algorithm, lacking temporal metadata, attributes
to direct causal links in the wrong direction.

\subsection{Constrained Analysis: Model Year as Exogenous}

\begin{figure}[t]
\centering
% ============================================================
% Figure 1: Default BOSS+FFML DAG (no background knowledge)
% ============================================================
\begin{tikzpicture}

  \circnode{mpg}{90}{mpg}
  \circnode{cylinders}{45}{cylinders}
  \circnode{displacement}{0}{displacement}
  \circnode{horsepower}{-45}{horsepower}
  \circnode{weight}{-90}{weight}
  \circnode{acceleration}{-135}{acceleration}
  \circnode{modelyear}{180}{modelyear}
  \circnode{origin}{135}{origin}

  \draw[directed] (cylinders)    -- (horsepower);
  \draw[directed] (displacement) -- (acceleration);
  \draw[directed] (displacement) -- (cylinders);
  \draw[directed] (displacement) -- (horsepower);
  \draw[directed] (displacement) -- (weight);
  \draw[directed] (horsepower)   -- (acceleration);
  \draw[directed] (horsepower)   -- (mpg);
  \draw[directed] (horsepower)   -- (weight);
  \draw[directed] (mpg)          -- (modelyear);
  \draw[directed] (origin)       -- (displacement);
  \draw[directed] (weight)       -- (acceleration);
  \draw[directed] (weight)       -- (modelyear);
  \draw[directed] (weight)       -- (mpg);

  \node[font=\scriptsize\sffamily, align=center] at (0,0)
    {(a) No background knowledge\\ Score: $-561.51$};

\end{tikzpicture}
\quad
%
% ============================================================
% Figure 2: BOSS+FFML DAG with modelyear constrained exogenous
% ============================================================
\begin{tikzpicture}

  \circnode{mpg}{90}{mpg}
  \circnode{cylinders}{45}{cylinders}
  \circnode{displacement}{0}{displacement}
  \circnode{horsepower}{-45}{horsepower}
  \circnode{weight}{-90}{weight}
  \circnode{acceleration}{-135}{acceleration}
  \circnode{modelyear}{180}{modelyear}
  \circnode{origin}{135}{origin}

  \draw[directed] (displacement) -- (acceleration);
  \draw[directed] (displacement) -- (cylinders);
  \draw[directed] (displacement) -- (horsepower);
  \draw[directed] (horsepower)   -- (acceleration);
  \draw[directed] (modelyear)    -- (mpg);
  \draw[directed] (modelyear)    -- (weight);
  \draw[directed] (mpg)          -- (displacement);
  \draw[directed] (mpg)          -- (horsepower);
  \draw[directed] (mpg)          -- (weight);
  \draw[directed] (origin)       -- (displacement);
  \draw[directed] (origin)       -- (mpg);
  \draw[directed] (weight)       -- (acceleration);
  \draw[directed] (weight)       -- (displacement);

  \node[font=\scriptsize\sffamily, align=center] at (0,0)
    {(b) \texttt{modelyear} exogenous\\ Score: $-613.27$};

\end{tikzpicture}

\caption{DAGs returned by BOSS\,+\,FFML (DAG mode) on the Auto MPG
  dataset ($n=392$, one discrete variable: \texttt{origin}).
  All edges are directed because FFML is not score-equivalent and
  can orient edges within a Markov equivalence class.
  \emph{(a)}~No background knowledge; the higher-scoring graph
  ($-561.51$) describes physical mechanisms sensibly but orients
  \texttt{modelyear} as a consequence of \texttt{mpg} and
  \texttt{weight}, which is causally implausible.
  \emph{(b)}~\texttt{modelyear} constrained to be exogenous
  ($-613.27$); the lower score reflects the data's genuine fit to
  the temporal trend, but the constrained graph admits a coherent
  design-priority interpretation in which \texttt{mpg} is an
  engineering target and \texttt{weight}, \texttt{displacement}, and
  \texttt{horsepower} are downstream design decisions made to meet it.}
\label{fig:autompg}
\end{figure}
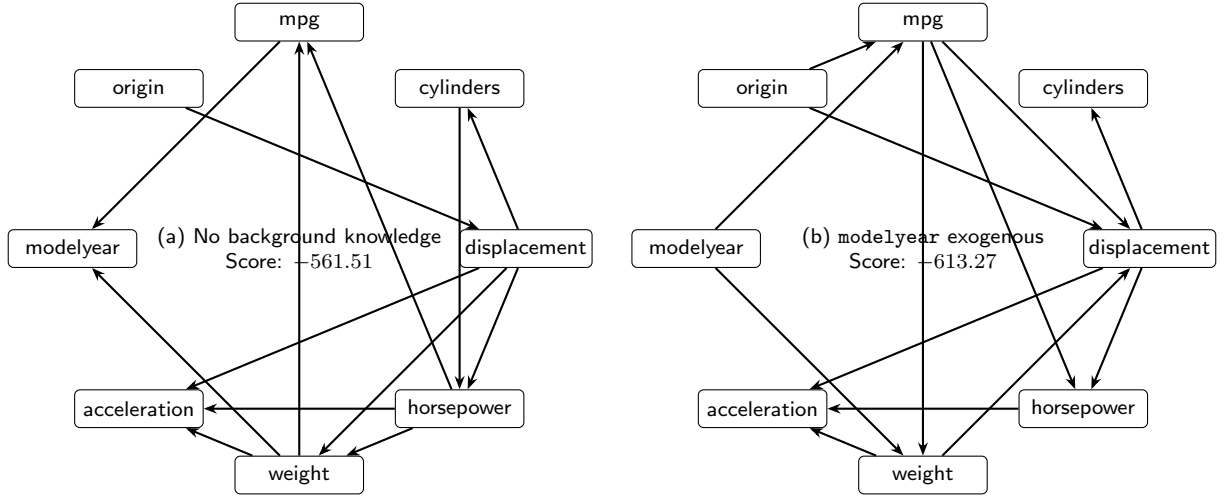

A user with domain knowledge can address this by placing
\texttt{modelyear} in a prior knowledge tier, constraining it to be
exogenous (i.e., it may be a cause of other variables but not caused
by them).
Figure~\ref{fig:autompg}(b) shows the resulting DAG (FFML score:
$-613.27$).

The constrained model scores lower than the unconstrained one
($-613.27$ vs.\ $-561.51$), which is the statistically honest result:
the data, without temporal metadata, genuinely fit a model in which
fuel economy and weight predict model year better than the reverse.
The value of the constraint is not statistical but interpretive ---
it enforces a causally coherent temporal ordering and reveals a
different and arguably more meaningful structure in the data.

The constrained graph admits a natural interpretation as a
\emph{design-priority model} rather than a physical-mechanism
model.\footnote{Thanks to Peter Spirtes for pointing out (in
conversation) this interpretation.}
In this reading, the arrows do not describe how one quantity physically
produces another, but rather how design targets propagate through
engineering decisions.
\texttt{modelyear}\,$\to$\,\texttt{mpg} captures the historical trend
of tightening fuel economy targets over time, driven by regulatory
pressure (particularly U.S.\ CAFE standards introduced in 1975) and
rising fuel prices.
\texttt{modelyear}\,$\to$\,\texttt{weight} reflects the contemporaneous
trend toward lighter vehicles.

The edges \texttt{mpg}\,$\to$\,\texttt{weight},
\texttt{mpg}\,$\to$\,\texttt{displacement}, and
\texttt{mpg}\,$\to$\,\texttt{horsepower} are the most striking feature
of the constrained graph, and they are precisely what the
design-priority interpretation predicts.
Engineers targeting a higher fuel economy specification choose vehicle
weight, engine displacement, and horsepower as the primary downstream
levers available to them: lighter, smaller, less powerful vehicles
achieve better fuel economy.
In this model, \texttt{mpg} is not a physical outcome but a design
target, and the mechanical parameters are consequences of meeting it.

\texttt{origin}\,$\to$\,\texttt{displacement} and
\texttt{origin}\,$\to$\,\texttt{mpg} reflect systematic differences in
engineering philosophy and regulatory environment across American,
European, and Japanese manufacturers, with the latter setting more
aggressive fuel economy targets and building correspondingly smaller,
lighter vehicles throughout this period.
\texttt{weight}\,$\to$\,\texttt{displacement} and
\texttt{weight}\,$\to$\,\texttt{acceleration} complete the picture:
once weight is determined by the mpg target, it in turn constrains
the displacement needed and governs acceleration performance.

The two graphs thus offer complementary readings of the same data.
The unconstrained graph (a) is the better statistical fit and describes
physical mechanisms in a largely sensible way, but misattributes the
direction of the temporal trend.
The constrained graph (b) scores lower but, under the
design-priority interpretation, tells a coherent story about how
fuel economy regulations and manufacturer philosophy shaped the
mechanical parameters of vehicles over the study period.
Neither graph is definitively ``correct''; the example illustrates
how background knowledge and causal interpretation interact with
data-driven search.

% ==============================================================
\section{Conclusion}
% ==============================================================

We have presented FFML, TRFF, and FFCI, three complementary methods
for nonlinear causal discovery in mixed continuous--discrete data.

FFML is a scalable approximation to the exact GP marginal likelihood,
replacing the $n \times n$ Gram matrix with a finite-dimensional RFF
representation while retaining the probabilistic interpretation and
automatic Occam-factor complexity penalty of the exact score.
It handles mixed parent sets through a product-kernel construction,
with a Kronecker feature-space path for small discrete parent sets
and a direct $n \times n$ Hadamard-product kernel path when the
discrete parent set is larger.
Bandwidth selection is stabilized via a grid search coupled to the
target variable, and feature coupling by target ensures stable score
differences across parent sets.

TRFF is a complementary BIC-style score using penalized Student-$t$
regression in random feature space. It achieves the highest precision
of any method in our experiments---both adjacency and arrowhead---at
the cost of lower recall, making it well-suited to settings where
false positive control is the primary concern. Together, FFML and
TRFF offer a precision--recall tradeoff that practitioners can
navigate based on their application priorities.

FFCI is a fast nonparametric CI test that featurizes each variable
individually, concatenating continuous RFF/ORF features with discrete
categorical features, and measures residual cross-covariance after
ridge residualization in feature space.
It supports four $p$-value approximation methods; the default gamma
approximation produces well-calibrated $p$-values under the null.
When run through the same PC-Max implementation, FFCI and RCIT exhibit
complementary precision--recall profiles: RCIT achieves higher
precision while FFCI achieves better recall and lower SHD, at
approximately twice the runtime.

Although FFML, TRFF, and FFCI share the same RFF/ORF kernel
approximation philosophy and ridge regularization strategy, they
differ architecturally: FFML builds a joint kernel over a parent set
for scoring; TRFF uses a concatenated design matrix with a BIC-style
penalty; FFCI featurizes variables individually for testing.
Together they form a practical and coherent toolkit for score-based,
constraint-based, and hybrid causal discovery at scale.

% ==============================================================
\section*{Revisions from Previous Version}
% ==============================================================

In a previous version of this paper (arXiv v1), the constraint-based
experiments reported results for PC\,+\,FFCI and PC\,+\,RCIT using
Tetrad's standard PC algorithm rather than PC-Max, due to a stale
parameter setting in the benchmarking script. The adjacency search
phase was unaffected, but the collider orientation step did not use
the PC-Max procedure described in Section~\ref{sec:experiments}.
The results reported in the current version use PC-Max throughout,
with all three constraint-based variants (FisherZ, FFCI, and RCIT)
running through the same Tetrad PC-Max implementation. The corrected
results show substantially improved arrowhead precision and recall for
PC-Max\,+\,FFCI relative to v1, and eliminate the catastrophic
orientation failures observed for PC\,+\,RCIT in v1, which were an
artifact of the PC implementation mismatch rather than a property of
the RCIT test itself.

In addition, data generation in the current version uses Tetrad's
\texttt{GeneralNoiseSimulation} class directly via a JPype bridge,
replacing an independent Python reimplementation used in v1. The noise
distribution is $\mathcal{N}(0, 0.3^2)$ rather than
$\tanh(\mathcal{N}(0,1))$ as stated in v1. All experimental results
have been updated accordingly.

% ==============================================================
\section*{Acknowledgements}
% ==============================================================

The author used a large language model (Claude, Anthropic) to assist with editing and
presentation of this document and takes full responsibility for its
content. The author was also supported by the U.S.\ Department of
Defense under Contract Number FA8702-15-D-0002 with Carnegie Mellon
University for the operation of the Software Engineering Institute.
The content of this paper is solely the responsibility of the author
and does not necessarily represent the official views of this funding
agency.

\bibliographystyle{plainnat}
\bibliography{refs}

@inproceedings{huang2018generalized,
  title={Generalized score functions for causal discovery},
  author={Huang, Biwei and Zhang, Kun and Lin, Yizhu and Sch{\"o}lkopf, Bernhard and Glymour, Clark},
  booktitle={Proceedings of the 24th ACM SIGKDD international conference on knowledge discovery \& data mining},
  pages={1551--1560},
  year={2018}
}

@article{strobl2019approximate,
  title={Approximate Kernel-Based Conditional Independence Tests
         for Fast Non-Parametric Causal Discovery},
  author={Strobl, Eric V. and Zhang, Kun and Visweswaran, Shyam},
  journal={Journal of Causal Inference},
  volume={7},
  number={1},
  year={2019},
  publisher={De Gruyter},
  doi={10.1515/jci-2018-0017}
}

@inproceedings{zhang2012kernel,
  title={Kernel-based Conditional Independence Test and Application
         in Causal Discovery},
  author={Zhang, Kun and Peters, Jonas and Janzing, Dominik and
          Sch{\"o}lkopf, Bernhard},
  booktitle={Proceedings of the 27th Conference on Uncertainty
             in Artificial Intelligence (UAI)},
  pages={804--813},
  year={2012}
}

@article{andrews2023fast,
  title={Fast scalable and accurate discovery of dags using the best order score search and grow shrink trees},
  author={Andrews, Bryan and Ramsey, Joseph and Sanchez Romero, Ruben and Camchong, Jazmin and Kummerfeld, Erich},
  journal={Advances in neural information processing systems},
  volume={36},
  pages={63945--63956},
  year={2023}
}

@inproceedings{rahimi2007random,
  title={Random Features for Large-Scale Kernel Machines},
  author={Rahimi, Ali and Recht, Benjamin},
  booktitle={Advances in Neural Information Processing Systems},
  volume={20},
  year={2007}
}

@inproceedings{yu2016orthogonal,
  title={Orthogonal Random Features},
  author={Yu, Felix and Suresh, Ananda Theertha and Choromanski,
          Krzysztof and Holtmann-Rice, Daniel and Kumar, Sanjiv},
  booktitle={Advances in Neural Information Processing Systems},
  volume={29},
  year={2016}
}

@article{ramsey2017million,
  title={A million variables and more: the fast greedy equivalence
         search algorithm for continuous variables and its extensions},
  author={Ramsey, Joseph and Glymour, Madelyn and Sanchez-Romero,
          Ruben and Glymour, Clark},
  journal={Journal of Machine Learning Research},
  volume={18},
  number={1},
  pages={1--47},
  year={2017}
}

@article{zheng2024causal,
  title={Causal-learn: Causal discovery in python},
  author={Zheng, Yujia and Huang, Biwei and Chen, Wei and Ramsey, Joseph and Gong, Mingming and Cai, Ruichu and Shimizu, Shohei and Spirtes, Peter and Zhang, Kun},
  journal={Journal of Machine Learning Research},
  volume={25},
  number={60},
  pages={1--8},
  year={2024}
}

@misc{auto_mpg_9,
  author       = {Quinlan, R.},
  title        = {{Auto MPG}},
  year         = {1993},
  howpublished = {UCI Machine Learning Repository},
  note         = {{DOI}: https://doi.org/10.24432/C5859H}
}

@inproceedings{ramsey2023py,
  title={Py-tetrad and rpy-tetrad: A new python interface with r support for tetrad causal search},
  author={Ramsey, Joseph and Andrews, Bryan},
  booktitle={Causal Analysis Workshop Series},
  pages={40--51},
  year={2023},
  organization={PMLR}
}

@article{chickering2002optimal,
  title={Optimal structure identification with greedy search},
  author={Chickering, David Maxwell},
  journal={Journal of machine learning research},
  volume={3},
  number={Nov},
  pages={507--554},
  year={2002}
}

@article{ramsey2016improving,
  title={Improving accuracy and scalability of the pc algorithm by maximizing p-value},
  author={Ramsey, Joseph},
  journal={arXiv preprint arXiv:1610.00378},
  year={2016}
}

@inproceedings{meek1995causal,
  author    = {Meek, Christopher},
  title     = {Causal inference and causal explanation with background knowledge},
  booktitle = {Proceedings of the Eleventh Conference on Uncertainty in Artificial Intelligence},
  series    = {UAI'95},
  pages     = {403--410},
  year      = {1995},
  publisher = {Morgan Kaufmann Publishers Inc.},
  address   = {San Francisco, CA, USA}
}

\end{document}